\documentclass{article}

\usepackage[final]{corl_2018} 

\usepackage{url}            
\usepackage{epsfig}
\usepackage{booktabs}       
\usepackage{amsfonts}       
\usepackage{nicefrac}       
\usepackage{microtype}      
\usepackage{bbm}      
\usepackage{graphicx}
\usepackage{amsmath,bm}
\usepackage{color}
\usepackage{caption}
\usepackage[font=footnotesize,labelfont=bf]{caption}
\usepackage[font=scriptsize,labelfont=sf,list=true]{subcaption}
\usepackage{times}
\usepackage{epsfig}
\usepackage{wrapfig,lipsum,booktabs}
\usepackage{graphicx}
\usepackage{amsmath}
\usepackage{amssymb}
\usepackage{csquotes}
\usepackage[british]{babel}
\usepackage{multirow}
\usepackage{color}
\usepackage{xcolor}
\usepackage{bm}
\usepackage{xcolor}
\usepackage{subcaption}
\usepackage{booktabs}
\usepackage{enumitem}
\usepackage{lipsum}
\usepackage{algpseudocode,algorithm,algorithmicx}
\usepackage{comment}

\setlist[itemize]{leftmargin=20 pt}
\usepackage[sans]{dsfont} 
\usepackage{wrapfig}
\usepackage{enumitem}
\usepackage[olditem,oldenum]{paralist}
\usepackage{soul}

\usepackage{multirow}
\usepackage{rotating}
\usepackage{booktabs}

\usepackage{alltt}
\usepackage{listings}

\usepackage{paralist}
\usepackage{mysymbols}
\belowdisplayskip \abovedisplayskip

\newlength{\sectionReduceTop}
\newlength{\sectionReduceBot}
\newlength{\subsectionReduceTop}
\newlength{\subsectionReduceBot}
\newlength{\abstractReduceTop}
\newlength{\abstractReduceBot}
\newlength{\captionReduceTop}
\newlength{\captionReduceBot}
\newlength{\subsubsectionReduceTop}
\newlength{\subsubsectionReduceBot}

\newlength{\eqnReduceTop}
\newlength{\eqnReduceBot}

\newlength{\horSkip}
\newlength{\verSkip}

\newlength{\figureHeight}
\newcommand{\vissys}{$\mathcal{V}$\xspace}
\newcommand{\qgen}{$\mathcal{Q}$\xspace}
\newcommand{\oraclev}{$\mathcal{O}$\xspace}
\newcommand{\adig}{$\mathcal{D}$\xspace}

\newcommand{\oracle}{Oracle\xspace}

\newcommand{\devi}[1]{\textcolor{red}{#1}}

\newcommand{\jianwei}[1]{\textcolor{blue}{JY: #1}}

\newcommand{\TODO}[1]{\textbf{\textcolor{red}{TODO: #1}}}

\title{Visual Curiosity: Learning to Ask Questions \\ to Learn Visual Recognition}

\author{
    Jianwei Yang$^{1}$\thanks{Equal contribution} \hspace{0.5pc}
    Jiasen Lu$^{1*}$ \hspace{0.5pc}
    Stefan Lee$^1$ \hspace{0.5pc}
    Dhruv Batra$^{1,2}$ \hspace{0.5pc} 
    Devi Parikh$^{1,2}$ \\
    {\small $^1$Georgia Institute of Technology} \hspace{0.5pc}
    {\small $^2$Facebook AI Research}
}

\begin{document}
\maketitle


\begin{abstract}
In an open-world setting, it is inevitable that an intelligent agent (\eg, a robot) will encounter visual objects, attributes or relationships it does not recognize. In this work, we develop an agent {empowered with \textit{visual curiosity}}, \ie the ability to ask questions to an \oracle (\eg, human) about the contents in images (\eg, \myquote{What is the object on the left side of the red cube?}) and build visual recognition model based on the answers received (\eg, \myquote{Cylinder}). In order to do this, the agent must (1) understand what it recognizes and what it does not, (2) formulate a valid, unambiguous and informative `language' query (a question) to ask the \oracle, (3) derive the parameters of visual classifiers from the \oracle response and (4) leverage the updated visual classifiers to ask more clarified questions.

Specifically, we propose a novel framework and formulate the learning of \textit{visual curiosity} as a reinforcement learning problem. In this framework, all components of our agent -- visual recognition module (to see), question generation policy (to ask), answer digestion module (to understand) and graph memory module (to memorize) -- are learned entirely end-to-end to maximize the reward derived from the scene graph obtained by the agent as a consequence of the dialog with the \oracle.

Importantly, the question generation policy is disentangled from the visual recognition system and specifics of the `environment' (scenes). Consequently, we demonstrate a sort of `double' generalization -- our question generation policy generalizes to new environments \emph{and} a new pair of eyes, \ie, new visual system. Specifically, an agent trained on one set of environments (scenes) and with one particular visual recognition system is able to ask intelligent questions about new scenes when paired with a new visual recognition system.

Trained on a synthetic dataset, our results show that our agent learns new visual concepts significantly faster than several heuristic baselines -- even when tested on synthetic environments with novel objects, as well as in a realistic environment.
\end{abstract}


\section{Introduction}
\label{sec:intro}
\vspace{-1mm}

As the various artificial intelligence sub-fields (vision, language, reasoning) mature, we are beginning to see ambitious multi-disciplinary tasks being undertaken -- at the intersection of vision-and-language (\eg image captioning~\cite{vinyals2015show, xu2015show, Lu2018Neural}, visual question answering~\cite{VQA, zhu2016visual7w}, visual dialog~\cite{visdial}), vision-and-navigation~\cite{gupta2017cognitive, zhu2017target}, and vision-language-and-navigation~\cite{anderson2017vision, embodiedqa, gordon2017iqa}. These tasks (and others) implicitly rely on the assumption that agent's visual recognition system is mature enough (\ie can recognize scenes, objects, their attributes, relationships, \etc) to support these higher-level AI tasks. 


However, in an open world, it is inevitable that the agent will encounter some new visual content (new scenes, objects, attributes) that it has never seen before. In such cases, it is natural to consider whether the agent can simply `ask' a human or an \oracle to identify the novel content and build visual classifiers on the fly. Note that this is a challenging task since the agent must (1) understand what it recognizes and what it does not, (2) formulate a valid, unambiguous and informative `language' query (a question) to ask the \oracle, (3) derive the parameters of visual classifiers from the \oracle response and (4) leverage the updated visual classifiers to ask more clarified questions.

\begin{figure}[t]
 \centering 
 \includegraphics[width=0.9\linewidth]{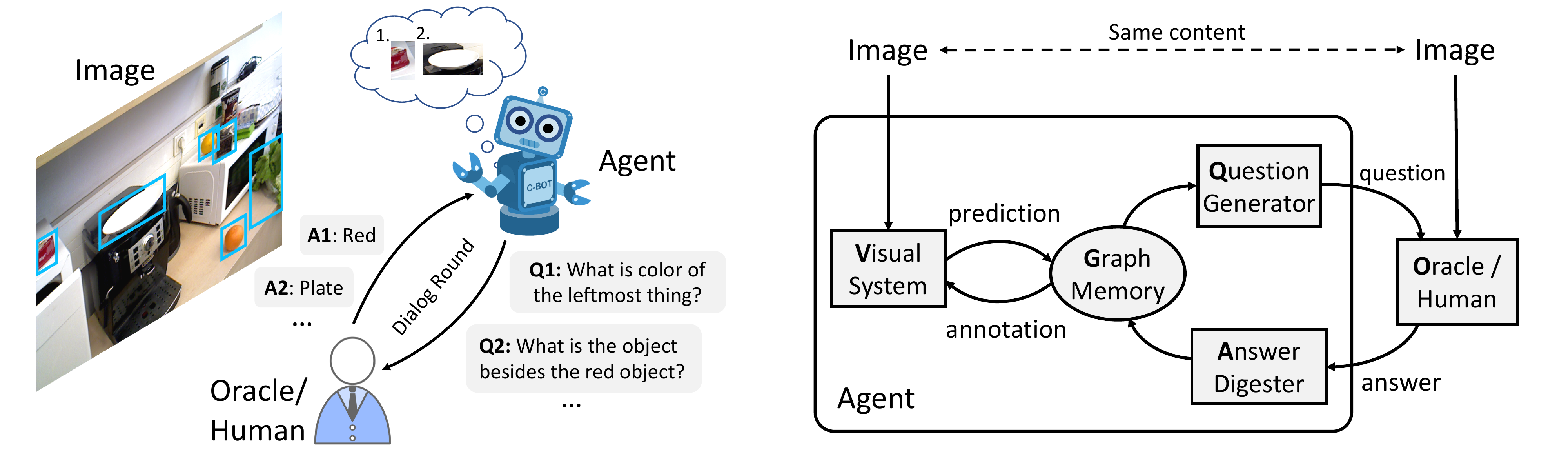}
 \caption{ Left: an example scenario where the agent learns to recognize objects through a dialog with an Oracle. Right: the proposed framework contains a visual recognition module (to see), question generation policy (to ask), answer digester (to understand) and graph memory module (to memorize).}
 \label{fig:teaser}
\end{figure}

Towards this goal, we develop an agent with the ability to ask questions about an image to an \oracle and build visual classifiers based on the answers received. We call this ability -- \textit{visual curiosity}. \figref{fig:teaser} left illustrates this setup. Given an image, the agent's visual system generates object proposals (or candidate bounding boxes). The agent is confident about labels of some candidate boxes (`orange\_fruit', `lettuce'), but does not recognize the content in others. It generates a question \myquote{What is the color of the leftmost object?}. The \oracle responds with the answer \myquote{red}, which the agent uses to update its `red' classifier. Furthermore, the agent uses the `red object' as a referent in future rounds of dialog to acquire labels of other objects (\myquote{What is the object besides the red object?}). 

One immediate question at this point may be 
-- what is the relationship of this setup to active learning \cite{kapoor2007active, li2013adaptive, settles2012active, vijayanarasimhan2014large}? 
A full discussion is available in \secref{sec:related_work}, but in short, our approach lies at the intersection of active learning and meta-learning -- \ie, instead of using a pre-specified active learning protocol, we \emph{learn to actively learn} \cite{bachman2017learning, contardo2017meta, fang2017learning}. Specifically, we formulate this task as a reinforcement learning problem and \emph{learn a 
policy} to ask questions to learn visual recognition. All components of our agent (illustrated in \figref{fig:teaser} right) -- visual recognition module (to see), question generation policy (to ask), answer digester (to understand) and graph memory module (to memorize) -- are learned entirely end-to-end to maximize the reward derived from the scene graph generated by the agent as a consequence of the dialog with the \oracle.

Importantly, the question generation policy is disentangled from the visual recognition system and specifics of the environment (scenes). Consequently, we demonstrate a sort of `double' generalization -- our question generation policy generalizes to new environments \emph{and} a new pair of eyes. 
Specifically, an agent trained on one set of environments (scenes) and with one particular visual 
recognition system is able to ask intelligent questions about new scenes 
when paired with a new visual recognition system 
(which may or may not recognize the same set of entities as the visual system during training). 

Our results show that our agent -- trained in a synthetic environment with a certain set of objects -- learns new visual concepts significantly faster than several heuristic baselines when deployed in a synthetic environment with novel objects as well as in a more realistic environment.

In order to make progress on this challenging problem, 
we make a number of simplifying assumptions that are described in detail in \secref{sec:approach} but highlighted here for completeness and full disclosure -- 
we use templated questions with slots that are filled by the agent, and model only simple geometric relationships between object proposals (right, left,  front, behind) that are trivial for the agent to extract from bounding box coordinates. Also, we assume the agent can localize objects in an image precisely. However, we believe the ideas and components of our work may generalize to more challenging scenarios in the future.

\section{Learning to Ask Questions}
\label{sec:approach}

As illustrated in \figref{fig:teaser}(b), there are four major components in our framework:

%
\begin{compactenum}[\hspace{5pt}1)]

\item \textbf{Visual System \vissys} that localizes image regions with high `objectness' (\ie generates object proposals) 
and predicts their categories and attributes. 

\item \textbf{Question Generator \qgen} that identifies an object proposal to inquire about and generates a question based on graph memory to ask the \oracle about its category or attribute;

\item \textbf{Answer Digester \adig} that uses the \oracle's answer to update the graph memory for training visual system \vissys to recognize the contents for future images; 

\item \textbf{Graph Memory} that is a semantic graph representation connecting the other three components. 

\end{compactenum}
%
\begin{figure}[t]
 \centering 
 \includegraphics[width=0.9\linewidth]{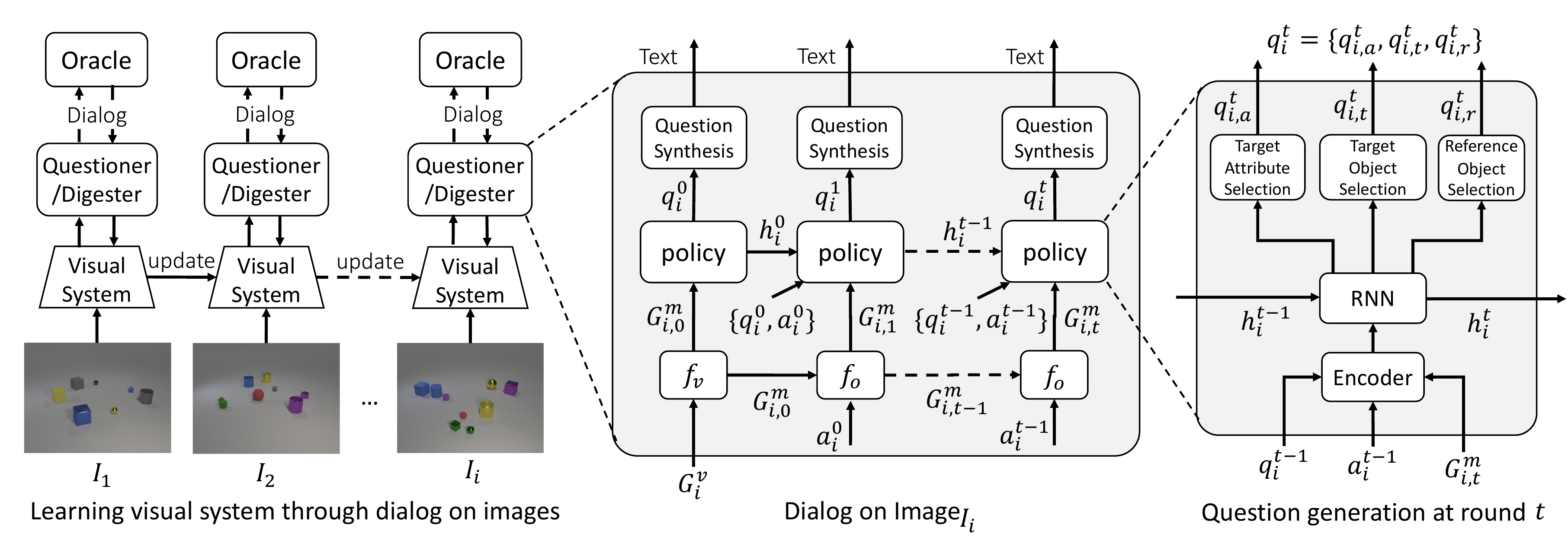}
 \caption{We simulate an agent observes a sequence of images, and interacts with the Oracle through dialogs to update its visual system (left). On each image, the agent asks a number of questions and gets responses from the Oracle (middle). For each question, the agent takes the history and the current graph memory as inputs and fills the question templates recurrently to compose a question (right).}
 \label{fig:model}
\end{figure}
%
\textbf{Graph Representation}. In our work, the agent's graph memory is the underlying data-structure connecting all other components; thus, we describe it first. 
It captures information about the image that the agent has gathered from the Oracle. For an image $I$, $G=(V, E)$ denotes a directed graph where the nodes $V$ correspond to the object proposals (with $\vert V \vert = K$), and edges $E$ correspond to the relationships between proposals. 
Let $\calA$ denote a set of visual attributes 
(\eg, object category, object shape) on object proposals. For an attribute concept $a \in \calA$, $n_{a}$ denotes the number of states for the concept $a$ (\eg concept \textit{color} can be `red', `blue', `green', etc). Let $\Delta^{n}$ be a $n$-simplex. Then, $\bm{p}^a \in \Delta^{n_a}$ denotes a probability distribution over attribute states for attribute concept $a$. Similarly, $n_{r}$ denotes the number of 
spatial relationships and $\bm{p}^r \in \Delta^{n_r}$ denotes a probability distribution over these relationships. Besides these distributions, each object proposal has a spatial location $\bm{l}$. 
We can then write the nodes of the graph as $V= \{(\bm{l}_k, \bm{p}^{a_1}_k, \dots, \bm{p}^{a_{\vert\calA\vert}}_k )\}_{k=1}^K$ and the edges as $E =\{p_{i\rightarrow j}^r \vert i,j \in [K], i\neq j\}$. In this work, the spatial relationships (left, right, behind, front) between object proposals are trivially recognizable from bounding box coordinates such that $p^r_{i\rightarrow j}$ are always delta functions. As such, we drop them from the graph notation for simplicity -- writing $G^m = \{(\bm{l}_k, \bm{p}^{a_1}_k, \dots, \bm{p}^{a_{\vert\calA\vert}}_k )\}_{k=1}^K$.
Besides the agent's graph memory, the agent also predicts a scene graph from an image using the visual recognition module $\mathcal{V}$. We denote this graph as $G^v = \{(\bm{l}_k, \bm{v}^{a_1}_k, \dots, \bm{v}^{a_{\vert\calA\vert}}_k )\}_{k=1}^K$ where $\bm{v}^a \in \Delta^{n_a}$. Likewise, the oracle \oraclev has an oracle graph $G^o = \{(\bm{l}_k, \bm{o}^{a_1}_k, \dots, \bm{o}^{a_{\vert\calA\vert}}_k )\}_{k=1}^{K^*}$ 
corresponding to the ground-truth scene graph with $K^*$ objects.  We use $G^m_i$, $G^v_i$ and $G^o_i$ to denote these graph representations for image $I_i$.

\textbf{Environment Setup}. To mimic the scenario of an agent traversing a novel environment while being instructed by a human about the world around it, we formalize our learning setup as a Markov Decision Process (MDP) over a series of image grounded dialogs. Specifically, an episode consists of multi-round dialogs about a sequence of $n$ images $I_1, ... I_n \in \mathcal{I}$. Our goal is to learn a good policy to discuss with the Oracle in turn one by one on these $n$ images so as to learn a good visual system to recognize objects and attributes. 
If successful, each of these dialogs with the Oracle produces valuable annotations on which to train the visual system which in turn produces a stronger foundation for subsequent dialogs.

\textbf{Rollout Process.} 
This environmental setup can be represented by a recurrent process as depicted in Fig.~\ref{fig:model} -- the agent initializes the graph memory using predictions from its visual system, the agent holds a dialog with the oracle to update this memory, and then the information gained over the dialog is used to update the visual system before this process is repeated for the next image. More formally, assume we have access to a question generation policy $\pi_q$, and a visual system $\mathcal{V}$. Presented with the image $I_i$, the agent first extracts the visual graph $G^v_i$ from the image with $\mathcal{V}$. Before beginning the dialog with the Oracle, the agent updates its initial graph memory $G^m_{i,0}$ based on $G^v_i$ through a bottom-up update function $f_v(G^m_{i,0}, G^v_i)$. Then, the agent engages in a $T$ round dialog with the oracle and maintains a sequence of graph memories $\{G^m_{i,0},...,G^m_{i, T}\}$ corresponding to its beliefs about the image $I_i$ at each round. At round $t$, the agent proposes a question $q_i^t$ using the policy $\pi_q$ based on the whole dialog history $\mathcal{H}_i^t = \{G^m_{i,0}, q^1_i, a^1_i, G^m_{i,1} ..., q^{t-1}_{i}, a^{t-1}_i, G^m_{i, t-1}\}$. The Oracle receives the question $q_i^t$ and generates an answer $a_i^t$ based on oracle graph $G^o_i$. Upon receiving the answer, the agent updates its graph memory using the top-down update function $f_o(G^m_{i,t-1}, a^t_i)$.

At the end of dialog on $I_i$, the agent uses the final graph memory $G^m_{i, T}$ along with the accumulated graph memories $\{G^m_{1, T}, ..., G^m_{i-1, T}\}$ to update the visual system \vissys before going to the next image $I_{i+1}$. This recurrent procedure on $n$ images is outlined in Alg.~\ref{algorithm1}. At the end of this process, the agent produces a trained visual system that can recognize the objects and attributes in images. We will elaborate the detail of each component in following section.

\subsection{Model}

We elaborate on each of the main components of our model in this section.

\begin{figure}[t]
  \centering
  \vspace{-10pt}\hspace{-5pt}
  \resizebox{0.85\columnwidth}{!}{
  \begin{minipage}{\linewidth}
    \begin{algorithm}[H]
      \caption{$Rollout(\{I_1, ..., I_n\}, T)$: MDP rollout process on $n$ images.}
      \textbf{Inputs}: Image sequence $\{I_1, ..., I_n\}$; Dialog budget $T$; Question generation policy $\pi_q$ \\
      \textbf{Outputs}: Visual system $\mathcal{V}$; Rewards $\{r_1^{1, T},..., r_n^{1, T}\}$ 
      \begin{algorithmic}[1]
      \State Initialize $G^o_{\{1,...,n\}}$ with ground truth, $G^m_{\{1,...,n\}}$ with uniform distribution
      \For  {$i \in [1 \cdots n]$ }
      \State $G^v_i \gets \mathcal{V}(I_i)$
      \Comment{Extract visual graph from $I_i$}
      \State $G^m_{i,0} \gets f_v(G^m_{i,0}, G^v_i)$
      \Comment{Initialize graph memory with visual graph}
      \For  {$t \in [1 \cdots T]$ }
      \State $q^t_i \gets \pi_q(\mathcal{H}_i^t)$
      \Comment{Generate question}
      \State $a^t_i \gets O(q^t_i, G^o_i)$
      \Comment{Oracle answers the question}
      \State $G^m_{i, t} \gets f_o(G^m_{i, t-1}, a_i^t)$
      \Comment{Update graph memory with answers}
      \EndFor
      \State Train $\mathcal{V}$ with $[G^m_{1, T} \cdots G^m_{i, T}]$
      \Comment{Train visual system with graph memories}
      \EndFor
      \end{algorithmic}
      \label{algorithm1}
    \end{algorithm}
  \end{minipage}}
\end{figure}

\textbf{Question Generator} \qgen. In order to produce queries to Oracle that are informative, 
the agent selects from a set of template questions -- filling in information from the graph memory. Inspired by \cite{johnson2017clevr}, each template is associated with a functional program that operates on the oracle scene graph to get the Oracle answer. For example, the question `What is the color of the metal object?' has a corresponding program: `$query\_color(unique(filter\_material(metal, scene)))$'.
%
%

Using these templates, the question generation is equivalent to selecting the objects and attributes about which to inquire. Specifically, the policy needs to determine which object attribute to ask about (\ie, target attribute), which object to ask about (\ie, target object), and if applicable which object to refer to (\ie, reference object). For instance, for the image in Fig.~\ref{fig:teaser}, the generated question may be ``What is the $<$\textit{white}$>$ $<$\textit{object}$>$ besides the $<$\textit{red object}$>$''. The target object and attribute is $<$\textit{object}$>$ and $<$\textit{white}$>$ respectively. The reference object is $<$\textit{red object}$>$.

We implement the question generation policy $\pi_q$ using a recurrent neural network (RNN). The memory provided by a recurrent policy is essential for the agent to know which questions have already been asked and whether they were meaningful or not according to the responses from Oracle (\ie referring to valid objects). As illustrated on the right side of Fig.~\ref{fig:model}, at each recurrent time step through the dialog, this policy takes as input the previous hidden state $h_i^{t-1}$, the previous round question $q_i^{t-1}$, its corresponding answer $a_i^{t-1}$, and the current graph memory $G^m_{i, t}$. Then, it outputs actions to select target attribute, target object, and reference objects. The selection of reference objects is handled by another recurrent process and may specify either none or one reference object.

\textbf{Oracle} \oraclev. Given the question from the agent, the Oracle answers the question by executing the functional program on the oracle graph $G^o_i$. However, the execution can fail in some cases. First, the question might be ambiguous. For example, the agent may ask `What is the color of the sphere?' when there are multiple spheres in the image. Second, the question might be invalid. For example, it is invalid if the agent asks the same question as above when there are no spheres in the image. As a result, the Oracle has three types of responses to the agent: 1) the answer to the question, 2) `ambiguous$\_$question' and 3) `invalid$\_$question'. If the Oracle responds with an answer, \eg, `red' to the agent, the agent's graph memory will be updated, otherwise it will stay the same.

\textbf{Updating the Graph Memory.} The graph memory is updated from the bottom-up (via visual system \vissys) and top-down (via answer digester \adig) with update functions $f_v$ and $f_o$, respectively:

\begin{compactitem}[\hspace{5pt}--]
\item \textbf{Bottom-Up} $\bm{f_v}$: For object $k$ and attribute $a$, its probability $\bm{p}_k^a$ is updated to a one-hot vector by setting its $\arg\max (\bm{v}_k^a)$-th entry to 1 and others to 0, if $\max(\bm{v}_k^a)>\tau_i$, where $\tau_i$ is a threshold that is annealed during the recurrent process, $\tau_i = \max(0.9, \exp(-i/n))$.

\item \textbf{Top-Down} $\bm{f_o}$: Suppose the agent asks about attribute $a$ for object $k$, and the answer is the $l$-th category for that attribute concept, then the agent will update its graph memory by setting the corresponding $l$-th entry in $\bm{p}_k^a$ to 1, and others to 0.

\end{compactitem}

\textbf{Reward.} A good question generator is one that asks meaningful questions to acquire knowledge about images from the Oracle. So we define the reward at each dialog round as:
\begin{equation}
\small
r_i^t = R(G^m_{i, t-1}, G^m_{i, t}, G^o_i) = S(G^m_{i,t}, G^o_i) - S(G^m_{i,t-1}, G^o_i)
\label{Eq:reward}
\end{equation}
where $S(\cdot)$ measures the similarity between the graph memory and the oracle graph. The reward is the difference in similarities between the current time step and the previous one. The purpose is to learn an agent that asks meaningful questions at \emph{each} time step so that it can recover as much information as possible within a budget of $T$ questions.

\subsection{Learning}
\begin{figure}[t]
  \centering
  \vspace{-10pt}\hspace{-10pt}
  \resizebox{0.9\columnwidth}{!}{
  \centering
  \begin{minipage}{\linewidth}
  	\setlength{\textfloatsep}{10pt}
    \begin{algorithm}[H]\footnotesize
\caption{Learning to Ask Question to Learn Visual Recognition.}
\textbf{Inputs}: Image sequence length $n$; Dialog budget $T$
\begin{algorithmic}[1]
\State Initialize parameters $\theta_{\pi}$ and $\theta_v$ for policy and visual system, respectively
\While {True}
\State Initialize parameters $\theta_v$
\Comment{Reset visual system at the beginning of episode}
\State $\bm{I} \gets \{I_1,...,I_n\} \sim \mathcal{E}$
\Comment{Sample $n$ images from an environment}
\State $\mathcal{V}, \{r_1^{1, T},..., r_n^{1, T}\} \gets Rollout(\bm{I}, T)$
\Comment{Rollout on $n$ images with Algorithm~\ref{algorithm1}}
\State Update $\theta_{\pi}$ based on $\{r_1^{1, T},..., r_n^{1, T}\}$ using Eq.~\eqref{Eq:policy}
\Comment{Train question generation policy}
\EndWhile
\end{algorithmic}
\label{algorithm2}
\end{algorithm}
  \end{minipage}}
\end{figure}

The overall learning algorithm for the agent is summarized in Alg.~\ref{algorithm2}. The policy $\pi_q$ is updated at the end of each episode while the visual system is updated multiple times during the inner $Rollout$.
Recall that our goal is to learn a strong question generation policy that can ask useful questions across varied environments and differently skilled visual systems.
To achieve this, we decouple the question policy from the visual system during training through two strategies: first, we introduce the semantic graph representation as an intermediate between perception and question generation; secondly, we reset the visual system to a random initialization at the beginning of each episode.
The visual system is trained inside the rollout process. At the end of dialogs on each image $I_i$, we append the graph memory to the history and use both for training the visual system. This training is a supervised learning task and the objective is:
\begin{equation}
\theta^*_v = \arg\min_{\theta_v} \sum_{I_i \in \bm{I}} \sum_{k = 1}^{K_i} \sum_{a \in \mathcal{A}} -\bm{p}_k^a \cdot \log(\bm{v}_k^a)
\end{equation}
where $K_i$ is the number of object proposals in $I_i$; $[\cdot]$ denotes the inner product operator between two vectors. The above objective is targeted to minimize the cross entropy between the prediction of visual system $\bm{v}_k^a$ and the graph memory $\bm{p}_k^a$, which is a one-hot vector as mentioned before. We use standard gradient descent methods to optimize this objective.

We train the question generation policy $\pi_q$ to recover as much information as possible from the Oracle in a limited budget, say $T$ dialog rounds. To succeed,  the agent must ask valid, unambiguous questions about uncertain object attributes.
To train the policy $\pi_q$ parametrized by $\theta_{\pi}$, we consider maximizing the expected reward gained by the policy over episodes under environment $\mathcal{E}$,
\begin{equation}
\theta^*_\pi = \arg\max_{\theta_\pi}~~J(\theta_\pi) ~=~ \arg\max_{\theta_\pi}~~%
\mathbb{E}_{\mathcal{I} \sim \mathcal{E}} %
\mathbb{E}_{\pi_q} \left[ \sum_{i = 1}^n \sum_{t=1}^{T} r_i^t(q_i^t \sim \pi_q(\mathcal{H}^t_i))\right].
\label{Eq:policy}
\end{equation}
%
In practice, we take a Monte Carlo estimate of this expected reward -- sampling a sequence of images
and questions throughout our dialogs -- and use advantage actor-critic \cite{mnih2016asynchronous} (A2C) to train our agent.




Details of the visual system, question templates, question generator can be found in Appendix.

\section{Related Work}
\label{sec:related_work}
\vspace{-2mm}
\noindent \textbf{Active Learning} addresses the problem of selecting samples from an unlabeled set to be labeled by some oracle \cite{kapoor2007active, li2013adaptive, settles2012active, vijayanarasimhan2014large}. Common selection criteria rely on heuristics, including entropy \cite{joshi2009multi}, expected model change \cite{settles2008multiple}, and boosting classifier margin \cite{deng2009imagenet}. 
Unlike traditional active learning, querying the oracle in our setting 
is not guaranteed to succeed; to gain a new label, agents must 
correctly refer to target objects when issuing queries to the Oracle. Further, our approach learns to collect labels efficeintly from end-to-end training rather than with predefined measures.

\noindent \textbf{Meta Active Learning.} Other recent work has also followed
this \emph{learning-to-active-learn} strategy \cite{bachman2017learning,contardo2017meta,fang2017learning},
training meta-learning models to select sets of instances to be labeled in order to maximize
performance of some target model trained on the selected set. As before, these models
have direct access to the oracle labels. Further, these meta-learners are tightly coupled
with their corresponding target model; being trained based on target model performance. In contrast, our approach is agnostic to the specific perception model.

%
%
%

\textbf{Learning by Asking Questions.} Mirsa \etal \cite{lbamisra17} present a learning-by-asking (LBA) framework for visual question answering (VQA). The main differences between our setting and LBA are two-fold: 1) We focus on learning a better visual system, not a better VQA model. Essentially, LBA is active learning (via language) for VQA, while our work is active learning (via language) to learn to see. 2) Our model decouples the visual system and question generation, which makes the learned question generator agnostic to different environments.  

\textbf{Teaching Robots via Language Interactions.} Previous work in human-robot interaction focus on agents learning new concepts from speaking with human operators \cite{thomason2017opportunistic, Yu2017LearningHT, skovcaj2016integrated, lutkebohle2009curious, sun2014learning, tellexll2013toward}. 
Tellex \etal \cite{tellexll2013toward} present a generalized grounding graph framework based on the linguistic coreference. However, there question generation policy and vision system are not designed to learn. L{\"u}tkebohle \etal \cite{lutkebohle2009curious} propose use language-interaction to solve ambiguity in the object references and for grasping commands. Thomason \etal \cite{thomason2017opportunistic} learn an active learning dialog policy for natural language grounding. 
Both \cite{Yu2017LearningHT} and \cite{skovcaj2016integrated} generate questions to continuous learn objects and visual properties.  
%
However, our goal is to learn a question generation policy that is distangled from the visual recognition system and specifics of the scenes, which enables both active learning and meta learning.
\section{Experiments}
\vspace{-2mm}
Recall that our goal is to learn visual curiosity, \ie, a question generation policy that can intelligently ask questions to an Oracle and in doing so acquire meaningful information to train a visual recognition system. A successful agent should work well not only in the setting it was trained, but also in new environments that contain partially or entirely novel attributes and with different visual systems ranging in levels of competency. Moreover, as the visual system is decoupled from question generation, the agent should generalize well to new visual domains, e.g.~from synthetic environments to realistic images. We evaluate our method for these qualities in the following experiments.

\subsection{Dataset}
\vspace{-2mm}
\begin{figure}[t]
 \centering
 \includegraphics[width=1\linewidth, scale=1]{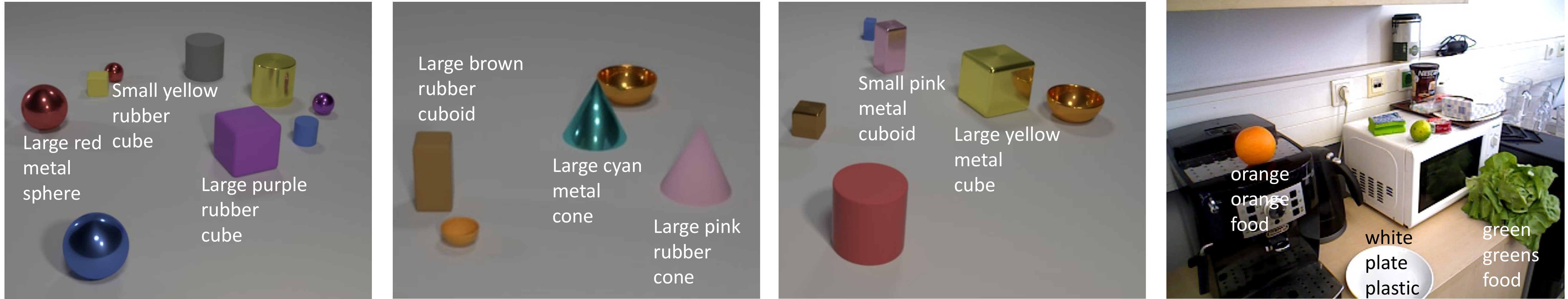}
 \caption{We use two types of datasets in our experiments. One is synthesized (left three columns) and one is a realistic dataset (right most). The synthesized one is further split to three sets, \textit{standard}, \textit{novel} and \textit{mixed}.}
 \label{fig:dataset}
\end{figure}

We evaluate our question generation policy in both synthesized and realistic environments. Exemplar images are shown in Fig.~\ref{fig:dataset}.
We generate the synthetic datasets using the same API as \cite{johnson2017clevr}. Each image contains 5 to 10 objects each with four different attribute types (\textit{shape}, \textit{color}, \textit{material}, and \textit{size}). We construct three different datasets to test generalization; specifically, we generate:

\begin{compactitem}[\hspace{5pt}--]
\item \textit{Standard} composed of objects from 3 shapes (cube, sphere, cylinder), 6 colors (gray, red, blue, green, yellow, purple), 2 materials (rubber, metal), and 2 sizes (large, small). 

\item \textit{Novel} consisting of objects from 3 novel shapes (cuboid, bowl, cone) and 4 new colors (pink, brown, cyan, orange) not present in \textit{Standard}; however, materials and sizes are the same. 
The goal is to check generalization to novel attribute values. 

\item \textit{Mixed} which contains objects from all 6 shapes, 10 colors, 2 materials and 2 sizes from both \textit{standard} and \textit{novel} splits. 
This is used to test the generalization ability on complex scenes where some attributes are known and others are not. 
\end{compactitem}







We synthesize 1800 images which we split  900/300/600 for train, val, and test respectively. The \textit{standard} train and val sets are used to train the agent policy, and the \textit{standard}, \textit{novel}, and \textit{mixed} test sets are used for evaluation. For the realistic dataset, we use the images and bounding boxes from the Autonomous Robot Indoor Dataset (ARID) \cite{arid}. It contains 153 objects from 51 categories. We further annotated each object with one of 6 different materials and one of 11 colors. The agent trained on the synthetic \textit{standard} split is also evaluated on this dataset. 


\subsection{Metrics and Baselines}
\vspace{-2mm}
For evaluation, we split each test set into 12 folds, each containing 50 images (i.e.~a single episode sequence). 
We run the learned agent on each fold and evaluate two metrics:
\begin{compactitem}[\hspace{3pt}--]
\item \textbf{Graph Recovery.} We measure the correctness of the agent's graph memory. This measures how informative the agent's questions were. We compute the graph memory's recall with respect to the ground truth as the percentage of correctly predicted attributes. We report the average graph recall across testing folds at dialog round K as R@K. We also report the area under this curve as AUC. 
\item \textbf{Visual Recognition}.  To evaluate if a better question generator leads to a better visual system, we measure how well a visual system performs after being trained through the agent's interactions with the Oracle on the test fold. To do so, we report the average graph recall of the visual system predictions on the remaining folds.
\end{compactitem}

%
%
%

We compare our proposed approach with three baselines:
\begin{compactitem}[\hspace{5pt}--]
\item \textbf{Random.} This agent randomly samples question to ask i.e~ it selects the target attribute, target object, and reference objects uniformly at random.
\item \textbf{Entropy.} 
An object/attribute with higher entropy (in the graph memory) is more likely to be chosen as the target. Likewise, objects with lower entropy are more likely to be references. 
\item \textbf{Entropy+Context.} The agent prefers to select uncertain (high entropy) object/attribute with reliable (low entropy) neighbors as reference objects. This way, the model prefers to ground questions on objects with low ambiguity.
\end{compactitem}
For comparisons, we make no changes other than replacing our approach with the above baselines.

\begin{table}[t]
\setlength{\tabcolsep}{3pt}
  \caption{Graph recovery performance (i.e., quality of questions asked) on the \textit{Standard}, \textit{Novel}, and \textit{Mixed} test sets for agents trained on \textit{Standard}.}
  \label{tab:result}
  \centering
  \resizebox{1\columnwidth}{!}{
  \begin{tabular}{l c c c c c c c c c c c c}
    \toprule
    \multicolumn{1}{c}{} & \multicolumn{4}{c}{\textit{Standard}}  & \multicolumn{4}{c}{\textit{Novel}} &  \multicolumn{4}{c}{\textit{Mixed}}\\    
   \cmidrule(r){2-5}
   \cmidrule(r){6-9}
   \cmidrule(r){10-13}
    Model  & R@10 & R@20 & R@50 & AUC  & R@10 & R@20 & R@50 & AUC  & R@10 & R@20 & R@50 & AUC\\
    \midrule	    
	\textbf{Random}    & 28.3 & 36.5 & 59.4 & 0.41 & 23.4 & 31.2 & 54.0 & 0.36 & 27.0 & 37.3 & 63.2 & 0.43 \\
	\textbf{Entropy} & 29.5 & 39.1 & 65.5 & 0.44 & 28.3 & 36.3 & 61.9 & 0.42 & 29.9 & 40.7 & 70.7 & 0.47 \\
    \textbf{Entropy+Context} & 38.0 & 52.5 & 67.1 & 0.52 & 35.2 & 46.5 & 59.5 & 0.46 & 38.5 & 49.9 & 66.4 & 0.52 \\
    \midrule
    \textbf{Our model} & \textbf{42.1} & \textbf{59.1} & \textbf{89.3} & \textbf{0.63} & \textbf{43.3} & \textbf{58.4} & \textbf{88.9} & \textbf{0.64} & \textbf{42.9} & \textbf{60.1} & \textbf{90.3} & \textbf{0.64} \\
    \textbf{Our model w/o} \vissys & 25.8 & 50.6 & 84.1 & 0.55 & 25.5 & 50.0 & 85.2 & 0.55 & 26.8 & 51.7 & 87.2 & 0.57 \\
   \bottomrule
  \end{tabular}}
\end{table}

\section{Results}
\vspace{-2mm}
Recall that we train on the \textit{standard} train set and evaluate on the \textit{standard}, \textit{novel} and \textit{mixed} test sets. 

\textbf{Questioner Graph Memory}. We first compare graph memory recovery for different models. As seen in Table.~\ref{tab:result}, our approach consistently outperforms the baseline models by a significant margin across all three test settings. Further, the performance between \textit{standard} and \textit{novel}/\textit{mixed} is similar, suggesting that our approach generalizes well to novel settings. 
The \textbf{Random} and \textbf{Entropy} baselines both struggle to propose unambiguous questions without the use of spatial context. 
The \textbf{Entropy+Context} model fairs better,
but falls off later when the hand-crafted strategy fails to find unambiguous reference objects. 
Our 
model 
steadily improves over the entire dialog and has apparently found a much better question asking strategy that generalizes well across different environments.

\textbf{Static Vision Ablation.} We also evaluate an ablated version of our model (\textbf{Ours w/o} \vissys) which never updates its visual system \vissys. This model must ask questions essentially from `scratch' without any bottom-up visual information. As shown in Table~\ref{tab:result}, the agent
starts dialogs with significantly lower graph similarity scores than our full model; however, as the dialog proceeds, this agent performs similarly. This highlights that the agent has learned to ask informative questions and not to simply rely on steadily improving the visual system.


\textbf{Visual System Performance}. We report the visual recognition accuracies in Fig.~\ref{fig:curves}(a-c). We take visual system checkpoints throughout the agent dialogs and evaluate them on the held out folds -- tracking the evolution of the visual system through the agent's interactions with the Oracle. We find our approach outperforms the baselines significantly in all settings. This is somewhat unsurprising as question generation and visual system learning are naturally synergistic -- with improvement of either leading to easier improvement in the other.



\subsection{Transferring to Realistic Environment}

Here, we apply the policy learned on the synthesized standard dataset to the realistic dataset. The episode length is also set to 50. 
In both situations, we observe significantly higher graph recalls for our model (86.2 R@50) than the baselines
(56.7, 66.1, 55.5, for random, entropy, and entropy+context respectively). 
More details are in Appendix. We also show the visual recognition curves in Fig.~\ref{fig:curves}(d). As we can see, the learned question generator can flexibly adapt to the realistic dataset and the learned visual system outperforms other baselines by a large margin. These results imply that our model could perhaps be deployed on a real embodied agent and learn a visual system from traversing in an environment with a guide.

\subsection{Inspecting the Question Generator}

\begin{figure}[t]
\centering 
\resizebox{1\columnwidth}{!}{
\begin{minipage}{.24\textwidth}
\includegraphics[width=1.0\linewidth, scale=1, trim=1 0 2 0,clip]{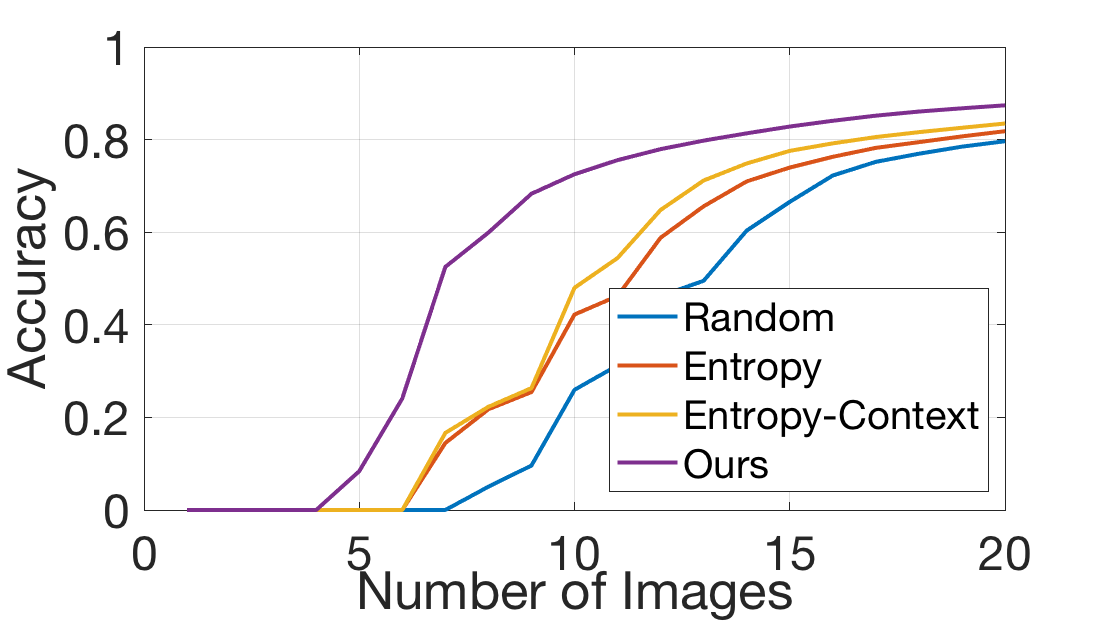} 
\subcaption{\scriptsize{\textit{Standard}}}
\end{minipage}
\begin{minipage}{.24\textwidth}
\includegraphics[width=1.0\linewidth, scale=1, trim=1 0 2 0,clip]{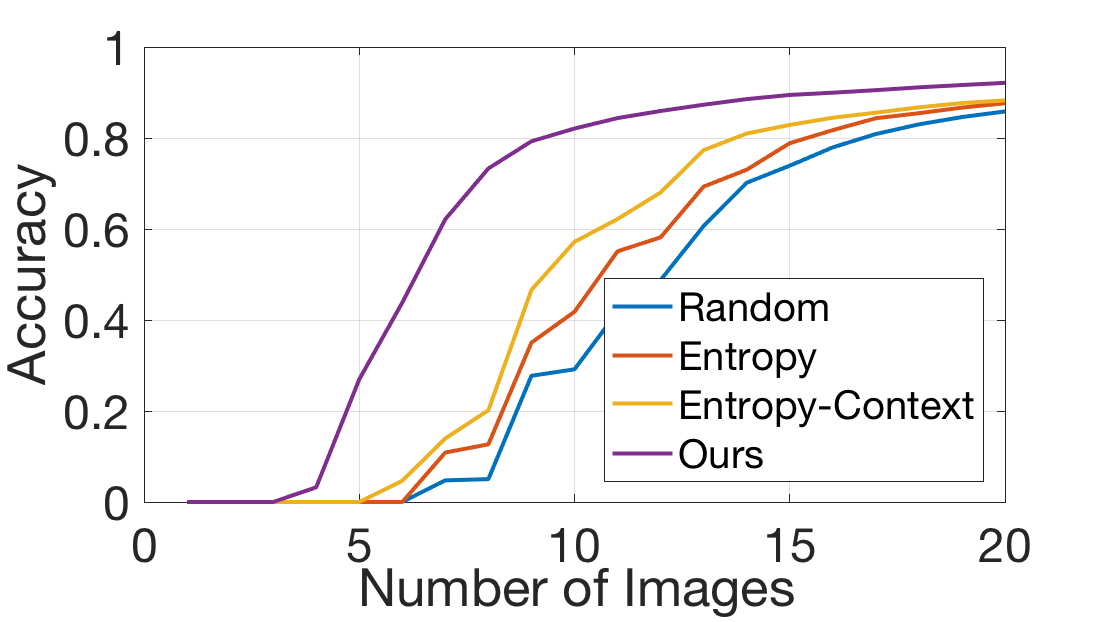}
\subcaption{\scriptsize{\textit{Novel}}}
\end{minipage}
\begin{minipage}{.24\textwidth}
\includegraphics[width=1.0\linewidth, scale=1, trim=1 0 2 0,clip]{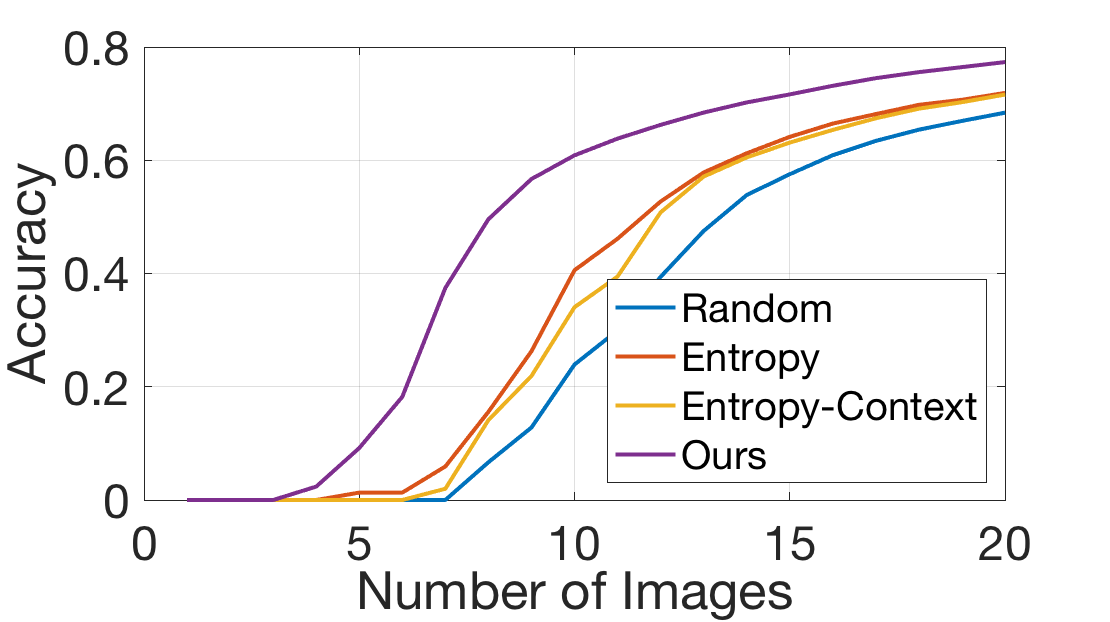}
\subcaption{\scriptsize{\textit{Mixed}}}
\end{minipage}
\begin{minipage}{.24\textwidth}
\includegraphics[width=1.0\linewidth, scale=1, trim=1 0 2 0,clip]{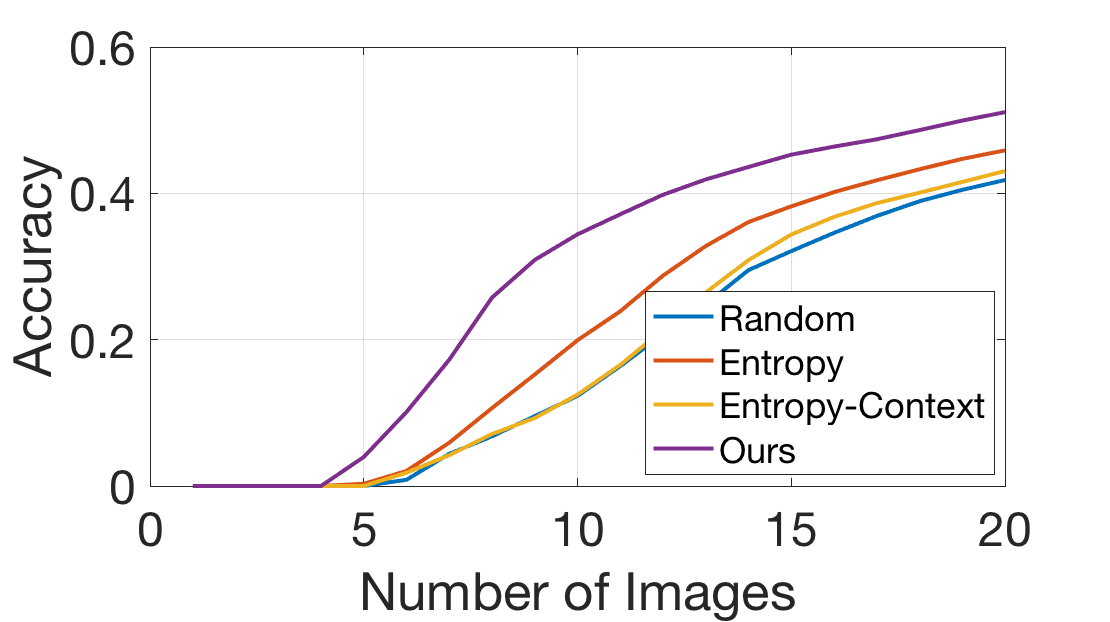}
\subcaption{\scriptsize{Realistic}}
\end{minipage}}

\centering
\resizebox{\columnwidth}{!}{
\begin{minipage}{.24\textwidth}
\includegraphics[width=1.0\linewidth, scale=1, trim=1 0 2 0,clip]{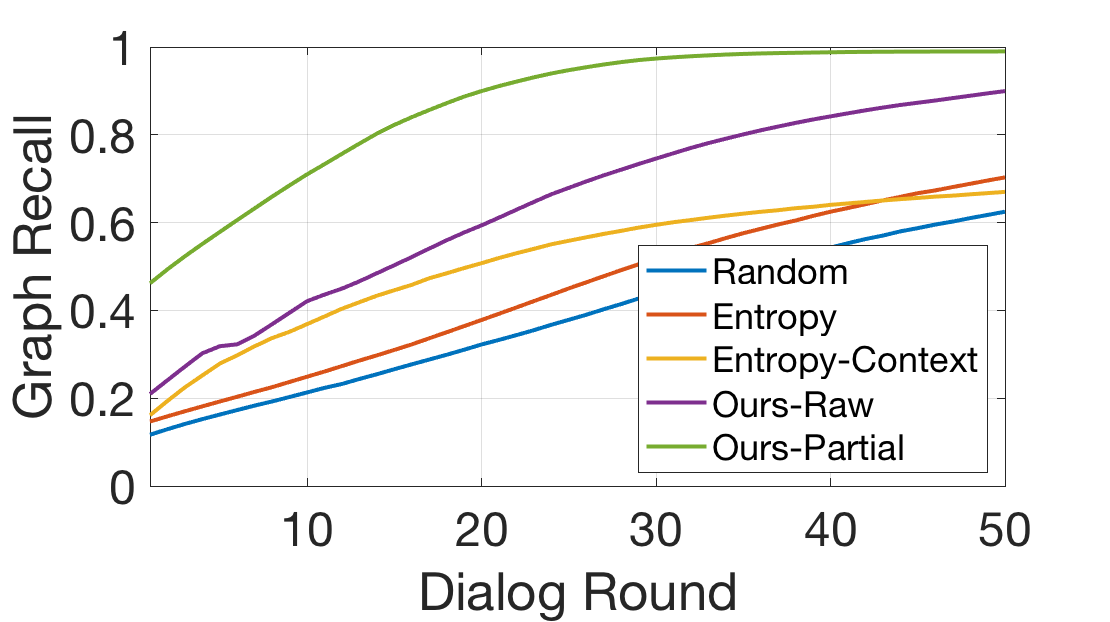}
\subcaption{}
\label{fig:partial_vs}
\end{minipage}
\begin{minipage}{.24\textwidth}
\includegraphics[width=1.0\linewidth, scale=1, trim=1 0 2 0,clip]{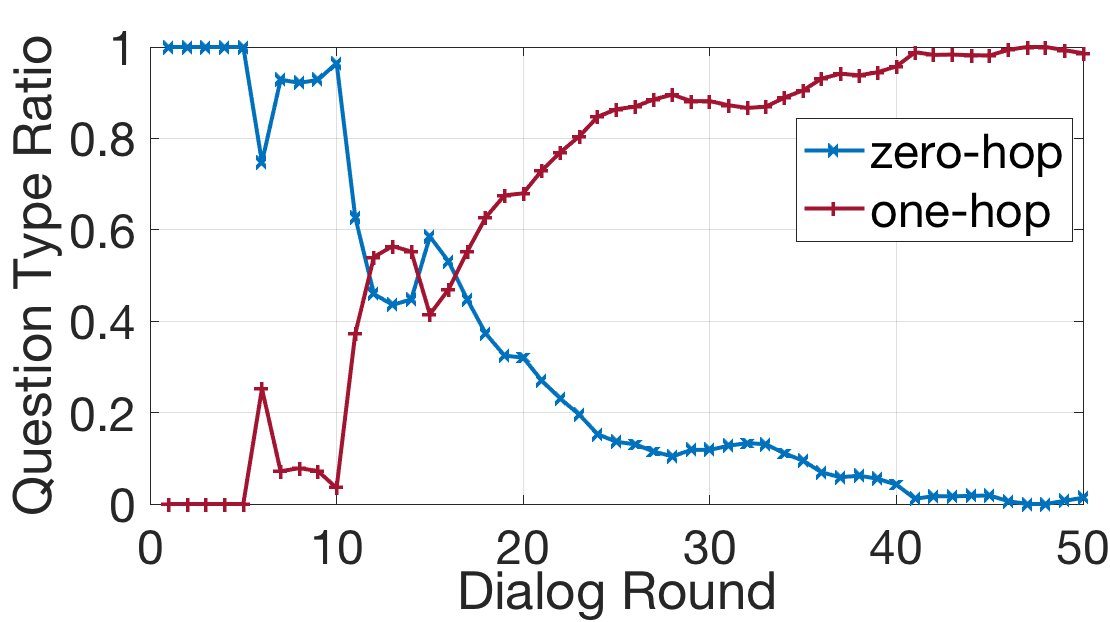}
\subcaption{}
\label{fig:ques_dist_round}
\end{minipage}
\begin{minipage}{.24\textwidth}
\includegraphics[width=1.0\linewidth, scale=1, trim=1 0 2 0,clip]{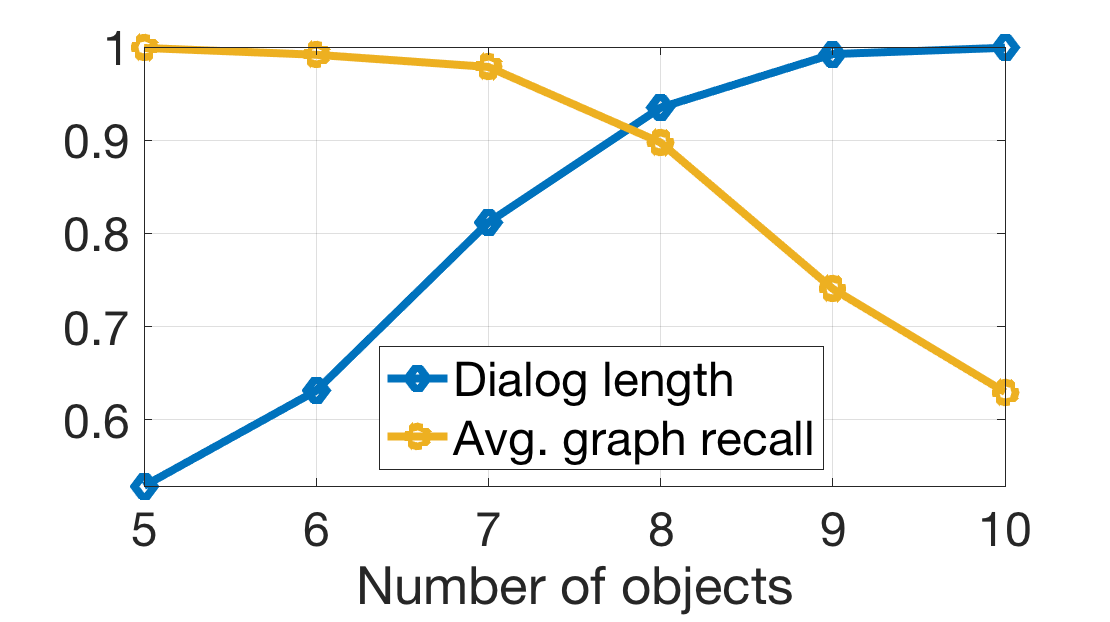}
\subcaption{}
\label{fig:recall_objnum}
\end{minipage}
\begin{minipage}{.24\textwidth}
\includegraphics[width=1.0\linewidth, scale=1, trim=1 0 2 0,clip]{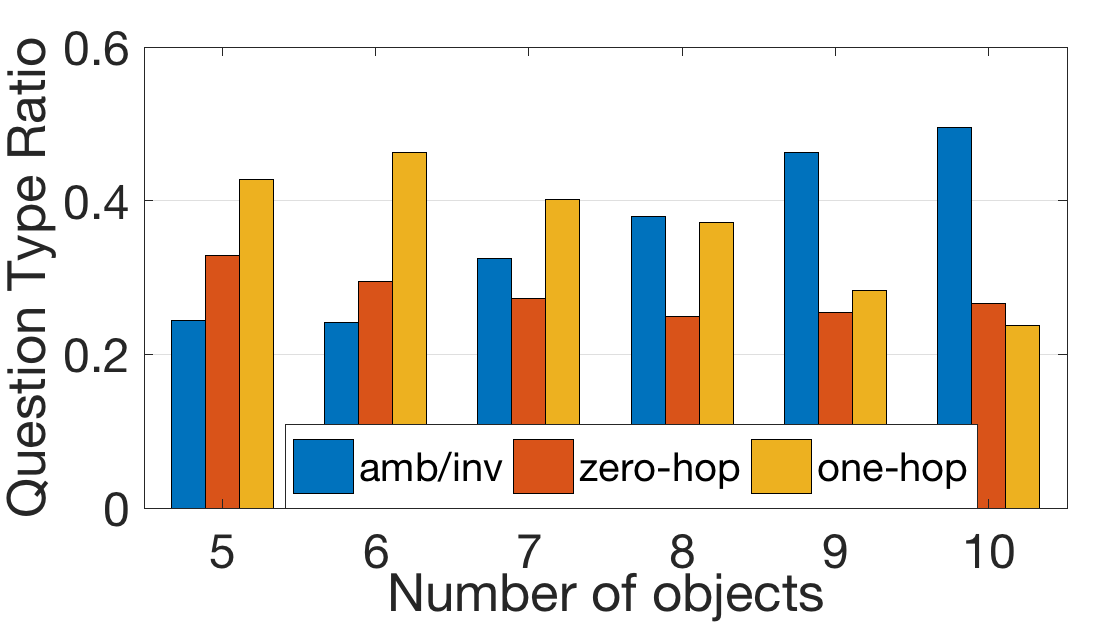}
\subcaption{}
\label{fig:ques_hist_objnum}
\end{minipage}}
\caption{Top Row: visual recognition accuracy curves against dialog round on different test sets. {Bottom Row: Inspecting different aspects of question generation.}}
\label{fig:curves}
\vspace{0mm}
\end{figure}

\textbf{Questioner starting with partially learned visual system}. {We investigate how the question generator behaves on the \textit{Mixture} set if its initial visual system can already recognize some of the attributes, \ie, those in the \textit{standard} set. In this case, the agent needs to ask about the remaining unknown attributes. In Fig.~\ref{fig:partial_vs}, we show average graph recalls on the \textit{Mixture} test set. We find the average graph recall for this ablation (Ours-Partial) starts from a much higher graph recall, and continues to increase to almost 1 -- indicating that the learned policy can also generalize well to partially trained visual systems. This is a promising result showing that the agent can leverage known visual concepts when learning about new ones and integrate additional visual systems seamlessly.}

\textbf{Question type against dialog round}. We run our question generator on all {\textit{Standard}} test images individually without updating the visual system. 
As shown in Fig.~\ref{fig:ques_dist_round}, it proposes more zero-hop questions at the beginning and then transitions to one-hop questions. This demonstrates that our model has learned an efficient strategy that asks questions about directly referable objects (e.g.~leftmost) first and then objects that require referring to other {(known)} objects. 

\textbf{Questioner behavior with varying number of objects}. We explore how the number of objects in images affect the question generation behavior. We separately evaluate the learned policy on images with varying number of objects. As shown in Fig.~\ref{fig:recall_objnum}, the average graph recall on images decreases and the relative dialog length (divided by maximal length 50, so can be plotted from 0 to 1)  increases when there are more objects in the images. In Fig.~\ref{fig:ques_hist_objnum}, we can see there are fewer unambiguous/valid questions when the number of objects increases -- implying that greater numbers of objects increases the difficulty for the questioner. However, our model can still perform well. As shown in Fig.~\ref{fig:curves}(g), our approach still achieves an average graph recall of 89.75 with 8 objects present.



\section{Conclusion}
\vspace{-2mm}
In this paper, we introduce a new setting \textit{learning visual curiosity}, where an agent \emph{learns} to ask questions to learn visual recognition. This is a challenging task where the agent needs to understand what it recognizes in an image and formulate language queries to the Oracle that are both unambiguous and informative. We use a graph memory to decouple the visual system and question generator. As a result, we demonstrate ``double'' generalization -- we show that the learnt policy to ask informative questions generalizes to new environments as well as to a new visual system. We experimentally demonstrate that a policy learnt on a synthetic set of objects generalizes to novel objects, to mixture of novel objects and attributes, as well as to a realistic dataset -- significantly better than strong baselines. This ability to learn about new objects and attributes by interacting with an Oracle is key to agents that operate in realistic open world settings.

\small{\textbf{Acknowledgements} }
\small{
This work was supported in part by NSF, AFRL, DARPA, Siemens, Google, Amazon, ONR YIPs and ONR Grants N00014-16-1-\{2713,2793\}. The views and conclusions contained herein are those of the authors and should not be interpreted as necessarily representing the official policies or endorsements, either expressed or implied, of the U.S. Government, or any sponsor.}

{\footnotesize
\bibliography{strings,egref}
}

\newpage
\section{Appendix}

\subsection{Question Generator}


As mentioned in Sec.~\ref{sec:approach}, our question generation policy $\pi_q$ is implemented using a recurrent neural network (RNN). The memory provided by a recurrent policy is essential for the agent to know which questions have already been asked and whether they were meaningful or not according to the responses from Oracle. As illustrated in Fig.~\ref{fig:model} right side, it consists of two components, target selection policy and reference selection policy. The first one determines which object and attribute to ask about. The second one determines whether to use a reference and which one to use if needed. These two policies share the low-level representations. Hence, we first elaborate the representation we use.

\textbf{Representation}. At the $t$-th dialog round on $I_i$, the question generator takes the question $q_{i}^{t-1}$ and answer $a_{i}^{t-1}$ from last round, and graph memory $G^m = \{(\bm{l}_k, \bm{p}^{a_1}_k, \dots, \bm{p}^{a_{\vert\calA\vert}}_k )\}_{k=1}^K$ as the input. Based on these three inputs, we compute:
\begin{itemize}
\item \textbf{Entropy of Graph Memory}: For object $k$ and its attribute $a$, we compute $e^a_k = Entropy(\bm{p}^a_k)$. For the whole scene graph memory, we obtain the entropy tensor $K \times |\mathcal{A}| \times 1$;
\item \textbf{Location Embedding}. For each object, we normalize its bounding box location $\bm{l}_k$ with image size and then use a two-layer MLP ($4-4-2$) to embed it to two dimensions. For all $K$ objects, the dimension is $K \times 2$. Afterward, we duplicate it for all attribute concepts, and thus obtain a tensor $K \times |\mathcal{A}| \times 2$;
\item \textbf{Target at last round}: For each of $K$ objects, we use one-hot tensor to encode which target object and which attribute the agent pointed to at last dialog round. Hence, the dimension is $K \times |\mathcal{A}| \times 1$.
\item \textbf{Reference at last round}: We use another one-hot vector to encode which reference object the agent pointed, and the dimension is $K \times 1$. We use another one-hot vector $K \times 1$ to record whether the agent use a reference object or not. If the agent does not use reference object, then the vector becomes zero vector. Similarly, We combine them and duplicate it for all attribute concepts to $K \times |\mathcal{A}| \times 2$;
\item \textbf{Answer at last round}: We use one-hot vector to encode the answer from Oracle at last round. If the answer is valid, then we assign 1 to the target and reference object slots; otherwise 0. As a result, we obtain $K \times |\mathcal{A}| \times 1$ and $K \times 1$ for target and reference, respectively. Afterward, we duplicate $K \times 1$ reference vector to $K \times |\mathcal{A}| \times 1$ and concatenate it with target tensor to obtain $K \times |\mathcal{A}| \times 2$.
\end{itemize}

Combining all the above signals, the final input to our question generator policy network at $t$-th dialog round is $x_t \in \mathcal{R}^{K \times |\mathcal{A}| \times 8}$. In our dataset, the number of attribute concept is 4. We replace $|\mathcal{A}|$ with 4 in the following for clarity. Given $x_t \in \mathcal{R}^{K \times 4 \times 8}$, we first reshaped it to $K \times 32$, where each row encode the graph memory and history for one object. Then we vectorize $x_t$ to $K$ vectors and feed them as a batch to a LSTM, obtaining new features $x^p_t \in \mathcal{R}^{K \times 64}$ by
\begin{equation}
h_t, c_t = lstm(h_{t-1}, c_{t-1}, x_t); \ x^p_t = h_t
\end{equation}
where $h_{t-1}$ and $c_{t-1}$ are the hidden state and cell memory from the lstm network at dialog round $(t-1)$. This $x^p_t$ from the hidden state in lstm will be used in both target and reference policy.

\textbf{Target policy}. It is aimed at pointing the right target object and attribute concept to ask about. This can be completed by directly pointing one of $K \times 4$ slots. To propose meaningful target and reference objects, the context is important. In our work, we exploit graph convolutional layers \cite{kipf2016semi} to pass the context information across different objects. Specifically, the GCN layer has the following typical formulation:
\begin{equation}
\bm{z}_{i}^{(l+1)} = \sigma \left( \sum_{j \in \mathcal{N}(i)} \alpha_{ij} W \bm{z}_{j}^{(l)} \right)
\label{Eq:gcn_orig}
\end{equation}
where $\mathcal{N}(i)$ is the neighbors of node $i$; $W$ is a learnable projection matrix; $\alpha_{ij}$ is the affinity between node $i$ and $j$. In our model, we compute the affinity between two object nodes based on the spatial distance:
\begin{equation}
\alpha_{ij} = \alpha_{ji} = \exp\left(-\frac{d(i, j)}{d_{max}}\right)
\end{equation}
where $d_{max}$ is the maximal distance in all object pairs. Given this affinity matrix, we first reshape $x^p_t$ to $(K \times 4) \times 16$ and pass the above tensor through two graph convolutional layers to obtain $x^{tar}_t \in \mathcal{R}^{(K \times 4) \times 16}$. Then we pass $x^{tar}_t$ through two-layer MLP (16-16-1) to obtain $(K \times 4)$ scores, and further a softmax layer to obtain a probability distribution $\bm{p}^{tar}_t = Softmax(mlp(x^{tar}_t))$. Besides the head for action, we have another head to compute the value. We simply perform average pooling for $x^{tar}_t)$ and also pass it to two-layer MLP to obtain the value for each of the object nodes. At end, to select the target object and attribute, we use an epsilon greedy sampling ($\epsilon=0.1$) strategy to choose one entry during training and choose the maximal one during testing.

\textbf{Reference policy}. It is aimed at determining whether to use reference object and which one if needed. It also takes $x^p_t$ as input. To select the right reference, this policy needs to know which target object is selected. Suspect the $k$-th object is selected as the target, we take the corresponding $k$-th $1 \times 64$ vector, and replicate it, which is then concatenated with the remaining to obtain $x^{ref}_t \in \mathcal{R}^{(K-1) \times 128}$. To determine which object to select as the reference, we also use two graph convolutional layers to update $x^{ref}_t$ to 64 dimension. Then, output is sent to a two-layer MLP (64-32-1) to obtain the $K-1$ dimensional scores over all candidate reference objects. Similar to the way used in target policy, the reference object is selected based epsilon greedy during training and the entry with maximal score during testing. Meanwhile, $x_t^{ref}$ is fed into another two graph convolutional layers to 64-d, which are then average pooled to obtain a single 64-d vector. This vector is then sent to a two-layer MLP (64-32-1) to predict whether or not to use reference object. For both selecting reference and determining whether to use reference, we compute the value using $x^{ref}_t$ as input.

Based on the above policies, we can deterministically compose a question with the corresponding template and ask it to the Oracle. We will introduce the details on the question template we used in our experiments below.

\subsection{Question Templates}
In our question templates, we introduce four attribute concepts size ($<$Z$>$), color ($<$C$>$, material ($<$M$>$), shape ($<$S$>$). We use $<$R$>$ to depict the relation between two objects, which could be `left', `right', `front' and `behind'. Besides, we introduce the absolute spatial relationship $<$P$>$ to depict the spatial location of one object proposal to the whole image. According to its location, it can be `left-most', `right-most', `closest' `farthest' or `None' otherwise.  Further, we use $<$L$>$ to indicate whether the target object proposal is at the extreme location among all proposals that have the relationship $<$R$>$ to its reference. It can be `closest' if it is extreme, and `None' otherwise. For clarification, we show two exemplar question templates below:
\begin{itemize}
\item ``What shape is the $<$P$>$ $<$Z$>$ $<$C$>$ $<$M$>$ $<$S$>$?''
\item ``What size is $<$L$>$ $<$Z$>$ $<$C$>$ $<$M$>$ $<$S$>$ that is $<$R$>$ $<$Z$>$ $<$C$>$ $<$M$>$ $<$S$>$?''
\end{itemize}

\begin{figure}[t]
 \centering 
 \includegraphics[width=1\linewidth]{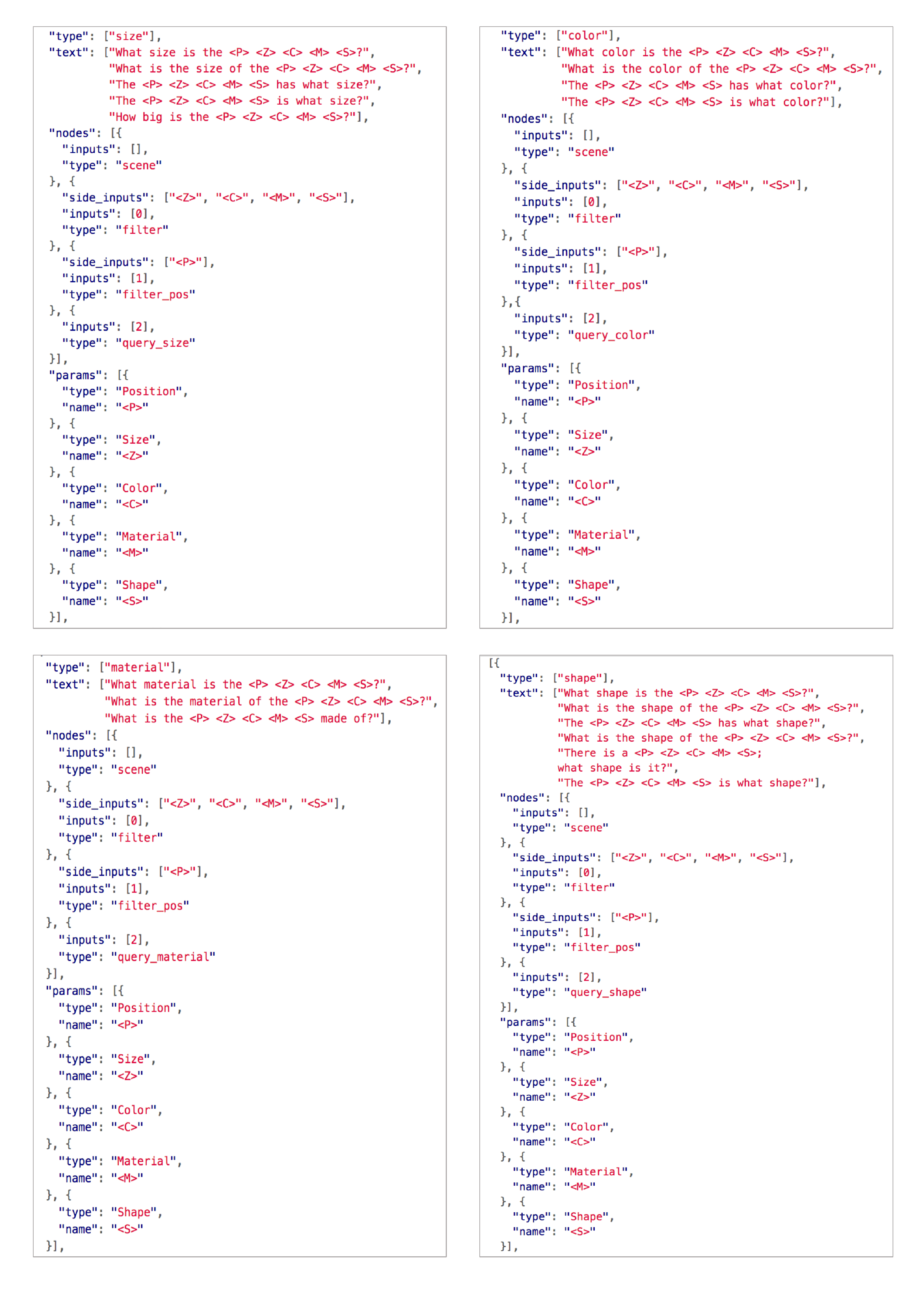}
 \caption{Zero-hop text and program templates on 4 attributes concepts (size, color, material, shape)}
 \label{fig:program0}	
\end{figure}

\begin{figure}[t]
 \centering 
 \includegraphics[width=1\linewidth]{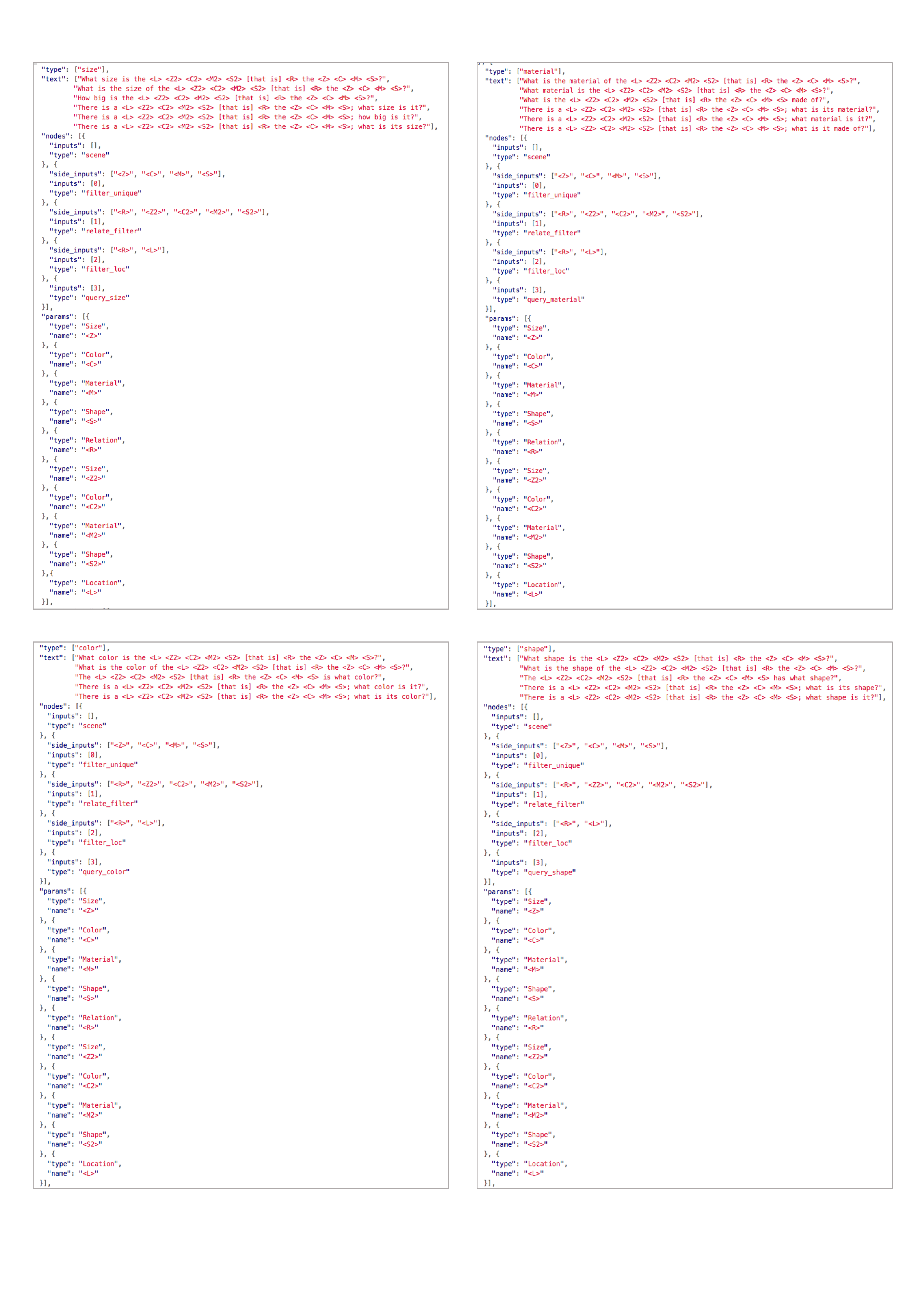}
 \caption{One-hop text and program templates on 4 attributes concepts (size, color, material, shape)}
 \label{fig:program1}	
\end{figure}

\qgen first points to the target object and reference object (if needed) and fill them into the above templates correspondingly. Based on the locations of target and reference objects, the relationship $<$R$>$, absolute location $<$P$>$ and relative location $<$L$>$ are manually inferred. Thus far, we can compose a unique questions which is then forward to Oracle side. Fig.~\ref{fig:program0} and Fig.~\ref{fig:program1} show the zero-hop and one-hop text and program templates on 4 attributes respectively. 

\subsection{Implementation Details}
We elaborate on the implementation details below:

\textbf{Visual system}. We use Faster-RCNN \cite{ren2015faster} in conjunction with a pre-trained VGG16 \cite{simonyan2014very} as the backbone of the visual system. We use the implemetnation open sourced in \cite{jjfaster2rcnn}. During training, the backbone is fixed, and we only learn the parameters for the four attribute classifiers (shape, color, size, material), which are two layer MLPs. We use the ground-truth bounding boxes on the agent side, since proposing object regions from the images is not our focus. In the future, we will try to use a region proposal network (RPN) to get the object proposals on the agent side. We are effectively assuming the the agent understands what constitutes an object, just not their names or attributes.


\textbf{Question templates}. We use zero-hop and one-hop question templates in our model. This is for two reasons: 1) they are enough to compose informative questions; 2) lower hop questions are more plausible to humans.
Two simplified exemplar templates are: 1) Zero-hop: ``What shape is the $<$\textit{Some Object}$>$?''; 2) ``What size is $<$\textit{One Object}$>$ that is $<$\textit{Spatial Relation}$>$ $<$\textit{Another Object}$>$?''

During training, the length of the sampled image sequences in an episode is set to 100 and we train the question generator over 200 episodes. The visual system is updated with 50 gradient descent steps during each update. We start the visual system update once it accumulates $5\times n_a$ annotations for attribute concept $a$. We use the Adam optimizer \cite{kingma2014adam} for the whole model. The learning rate starts from 1e-4 and decreases by a factor 0.99 after each image.

\subsection{Attribute Annotations for ARID}
\begin{figure}[t]
 \centering 
 \includegraphics[width=1\linewidth]{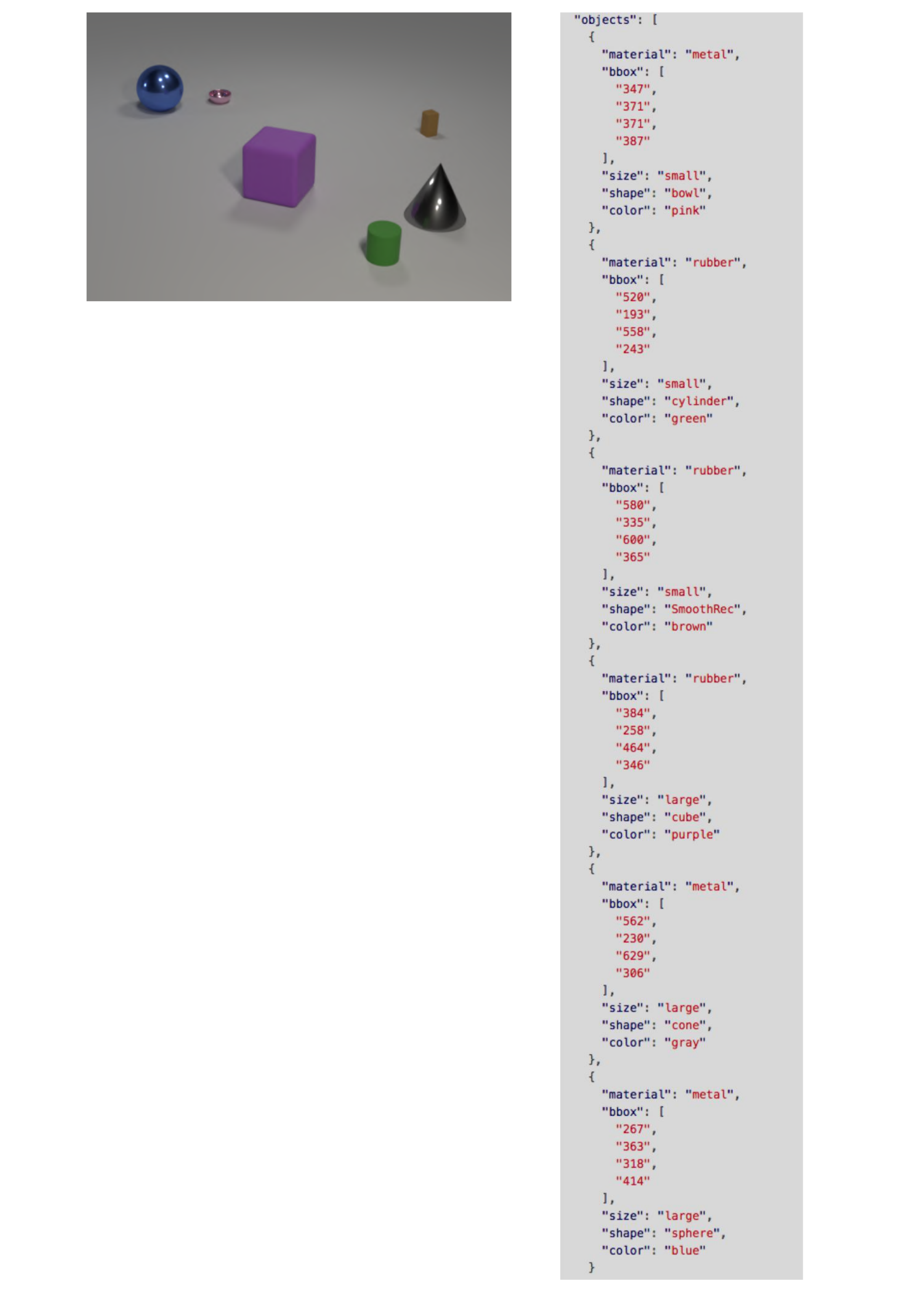}
 \caption{Example of an image and the associated scene
graph on \textit{mixture} synthesized dataset.}
 \label{fig:scene1}	
\end{figure}

\begin{figure}[t]
 \centering 
 \includegraphics[width=1\linewidth]{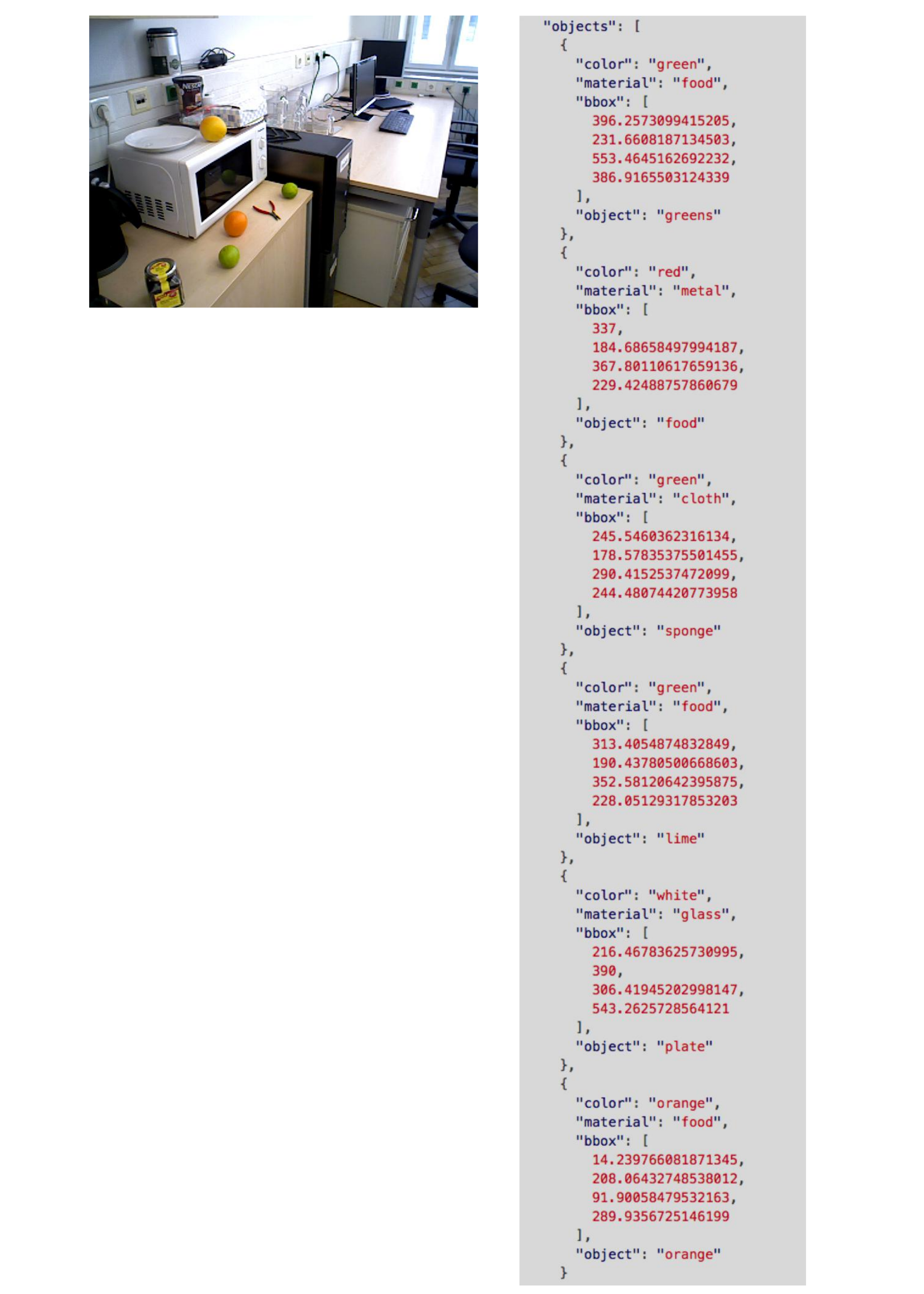}
 \caption{Example of an image and the associated scene
graph on ARID dataset.}
 \label{fig:scene2}	
\end{figure}

The attributes annotations for ARID \cite{arid} is shown in table~\ref{tab:arid}. We annotate three attribute concepts: object, color and material. Though the huge number of object categories than our synthetic dataset, our model trained on synthetic dataset generalizes very well to this new environment. Fig.~\ref{fig:scene1} and Fig.~\ref{fig:scene2} show the example of an image and the associated scene graph on \textit{mixture} synthesized dataset and ARID dataset.  

\begin{table}[!h]
\setlength{\tabcolsep}{3pt}
  \caption{Attributes annotations for ARID dataset, contains object, color and material.}
  \label{tab:arid}
  \centering
  \begin{tabular}{ll}
    \toprule
	\textbf{Object} & lightbulb, apple, bell, calculator, sponge, keyboard, marker, scissors, glue, lime, \\
          & flashlight, cell, lemon, instant, peach, toothpaste, bowl, rubber, camera, orange, \\
          & banana, plate, coffee, ball, mushroom, food, pear, pitcher, dry, kleenex, toothbrush \\ 
          & binder, notebook, garlic, cereal, pliers, comb, tomato, water, stapler, onion, greens, \\
          & potato, cap, shampoo, hand, soda 
          \\
          \midrule
    \textbf{Color} & blue, brown, purple, grey, yellow, mixed, pink, green, orange, black, white, silver, red\\
              \midrule
    \textbf{Materials} & cloth, food, metal, plastic, glass, paper\\ 
   \bottomrule
  \end{tabular}
\end{table}

\begin{figure}[!ht]
\begin{minipage}{0.5\textwidth}
\centering
\includegraphics[width=1.0\linewidth, scale=1]{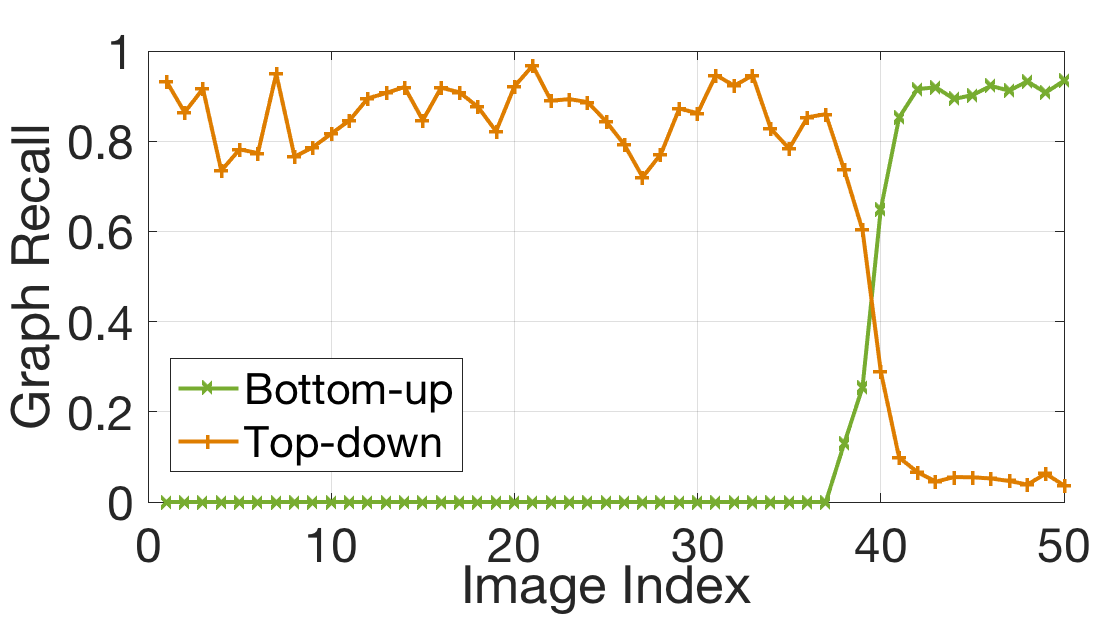}
\subcaption{}
\end{minipage}
\begin{minipage}{0.5\textwidth}
\centering
\includegraphics[width=1.0\linewidth, scale=1]{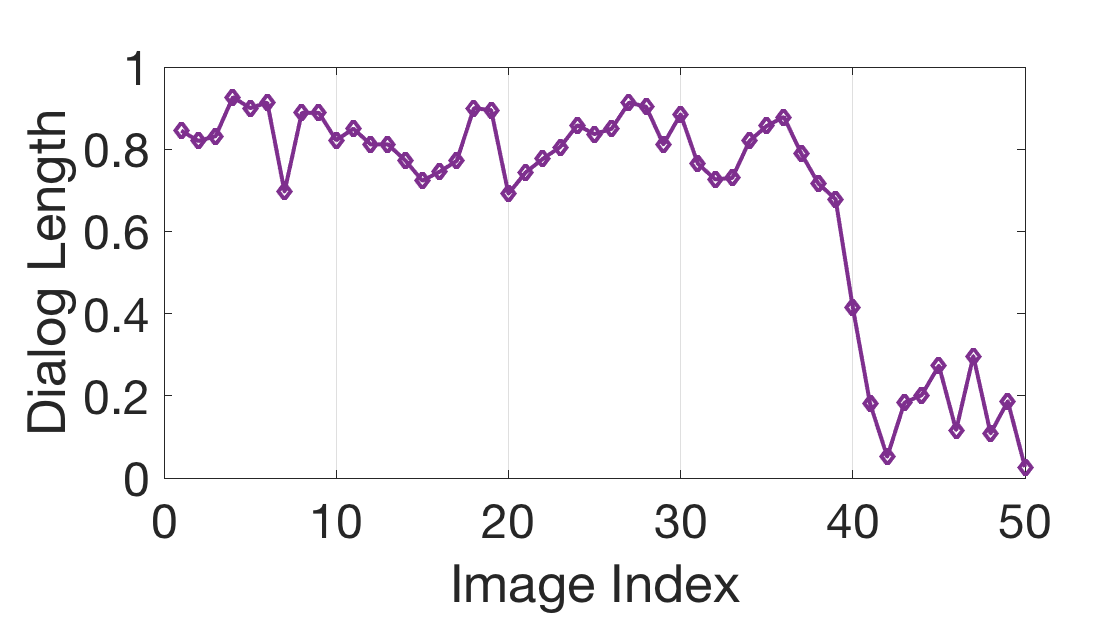}
\subcaption{}
\end{minipage}
\caption{Left: graph recall contributed bottom-up and top-down; Right: relative dialog length over time.}
\label{fig:inspect}
\end{figure}

\begin{table}[!ht]
\setlength{\tabcolsep}{3pt}
  \caption{Graph recall on realistic test set for policy trained on synthetic \textit{Standard} train set. }
  \label{tab:graph_recall}
  \centering
  \begin{tabular}{l c c c c c c c c c c c c}
    \toprule
    \multicolumn{1}{c}{} & \multicolumn{4}{c}{\textit{Realistic}} \\    
   \cmidrule(r){2-5}
   \cmidrule(r){6-9}
   \cmidrule(r){10-13}
    Model  & R@10 & R@20 & R@50 & AUC  \\
    \midrule	    
	\textbf{Random}  & 20.1 & 27.5 & 56.7 & 0.33  \\
	\textbf{Entropy} & 20.3 & 32.0 & 66.1 & 0.39  \\
    \textbf{Entropy+Context} & 30.2 & 39.9 & 55.5 & 0.41 \\
    \midrule
    \textbf{Ours model} & \textbf{35.6} & \textbf{53.4} & \textbf{86.2} & \textbf{0.59} \\
   \bottomrule
  \end{tabular}
\end{table}



\subsection{Graph recall from Bottom-up and Top-down}

Recall that the graph memory is updated both by the visual system as well as information from the oracle. We investigate the contributions of these {two factors} to the graph recovery over the dialogs. As shown in Fig.~\ref{fig:inspect}(a), as the dialog proceeds, the agent relies more on its visual system and less on interactions with Oracle. Since the graph memory is either updated bottom-up or top-down, we can easily measure their contributions by counting the number of entries updated by visual system and oracle in the graph memory. Also, as shown in Fig.~\ref{fig:inspect}(b), the number of dialog rounds drops. This is a plausible behavior since we {would not expect an intelligent agent} to keep asking questions repeatedly after multiple interactions with Oracle.

\subsection{Graph Recall on Realistic Environment}

As a supplement to the Section 5.1 in our main paper, in Table~\ref{tab:graph_recall}, we present the graph recalls for different methods on our collected realistic dataset. Clearly, our model outperforms all three baselines significantly. Though not being trained on the realistic environment, our questioner successfully propose meaningful questions to ask and get much higher graph recalls. Moreover, the numbers are comparable to those reported on synthetic test sets. \textbf{These numbers indicate that our model have a strong generalization ability across different environments.}

\subsection{Qualitative Results}

\begin{figure}[!h]
 \centering 
 \includegraphics[width=1\linewidth]{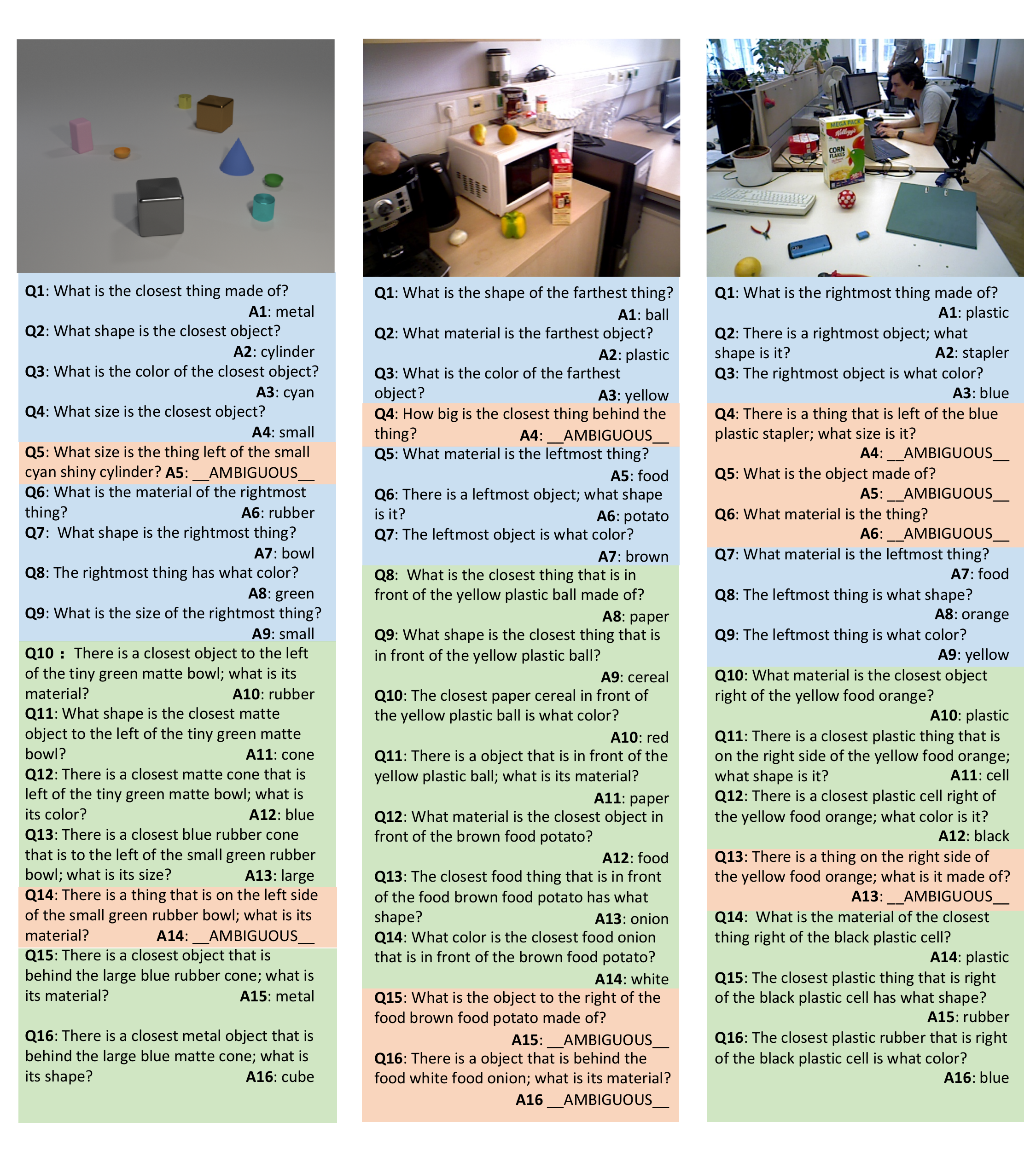}
 \caption{Dialogs with Oracle on \textit{Mixed} synthesized dataset (left) and ARID dataset (middle and right) based on the policy learned on \textit{normal} synthesized dataset. Questions in blue, green and red background corresponds to one-hop, two-hop and ambiguous questions respectively. }
 \label{fig:qualitative}
\end{figure}

In Fig.~\ref{fig:qualitative}, we show some qualitative results on both synthetic dataset and realistic dataset. Specifically, the questioner generation policy is trained on \textit{Standard} train set, and then applied to test sets. Here, we display the first 16 rounds of dialog with oracle on three images, which are from \textit{Mixed}, and ARID dataset. Our model learns to begin with zero-hop questions (blue), and followed with one-hop questions (green). Moreover, the learned question generation policy tends to repeatedly query one object until all the attributes are observed. When transferring to realistic environment, our method can successfully generalize to new objects and attributes, and ask meaningful questions. This verifies the effectiveness of our framework on disentangling the question generation policy and visual recognition system. We also find our model sometimes asks the ambiguous questions (red) which can not be answered by \oracle. The ambiguous questions can be either zero-hop question or one-hop question. When looking more closely, we find the ambiguous question is mainly caused by the unspecified target object, \eg, ``What size is the thing left of the small cyan shiny cylinder?'', ``There is a thing that ...''. However, by taking the current graph memory and histories, the agent can successfully get rid of this soon after a few dialog rounds with the Oracle.

\subsection{Limitations and Future Directions}

As we mentioned early, our work is an initial step towards learning visual curiosity. To start on this challenging task, we've made a number of simplifying assumptions that future work could soften. For instance, extending the model to more complicated visual scenarios where object proposal systems might be error-prone. In this case, the visual contents from the perspective of agent and Oracle are different from each other, which make the questions more ambiguous or confusing to the Oracle. One way to address this is empower the Oracle the visual curiosity ability as well, including answering clarifying questions to the agent. Another extension is considering richer sets of relationships between objects, and enable the agent to learn about relationships as well during the interaction with Oracle. Further, models could be extended to operate on non-templated dialog exchange, \ie, natural questions from agent and natural answers from Oracle. At last, in the current setting, we assume the number of the attribute concepts is given. However, incorporating with incremental learning to grow the attribute space over time would also be an interesting future direction to explore.

\end{document}